\documentclass[runningheads]{llncs}
\usepackage{amssymb}
\usepackage{mathtools}
\usepackage{enumitem}
\usepackage{multirow}
\usepackage{subcaption}
\usepackage{cite}
\usepackage[misc]{ifsym}
\usepackage[breaklinks=true,colorlinks,bookmarks=false]{hyperref}

\newcommand{\etal}{\textit{et al.}}

\begin{document}

%%%%%%%%% TITLE
\title{Adversarial Mask: Real-World Universal Adversarial Attack on Face Recognition Models}

\titlerunning{Adversarial Mask: Universal Adversarial Attack on Face Recognition Models}

\author{Alon Zolfi\inst{1}\orcidID{0000-0003-0270-1743}\Letter{} \and
Shai Avidan\inst{2} \and \\
Yuval Elovici\inst{1}\orcidID{[0000-0002-9641-128X} \and
Asaf Shabtai\inst{1}\orcidID{0000-0003-0630-4059}}
\authorrunning{A. Zolfi et al.}
\institute{Ben-Gurion University of the Negev, Israel \\
\email{zolfi@post.bgu.ac.il}, 
\email{\{elovici,shabtaia\}@bgu.ac.il}
\and
Tel Aviv University, Israel\\
\email{avidan@tauex.tau.ac.il}\\
}
\maketitle

\begin{abstract}

Deep learning-based facial recognition (FR) models have de\-monstrated state-of-the-art performance in the past few years, even when wearing protective medical face masks became commonplace during the COVID-19 pandemic.
Given the outstanding performance of these models, the machine learning research community has shown increasing interest in challenging their robustness.
Initially, researchers presented adversarial attacks in the digital domain, and later the attacks were transferred to the physical domain.
However, in many cases, attacks in the physical domain are conspicuous,
and thus may raise suspicion in real-world environments (e.g., airports).
In this paper, we propose \textit{Adversarial Mask}, a physical universal adversarial perturbation (UAP) against state-of-the-art FR models that is applied on face masks in the form of a carefully crafted pattern.
In our experiments, we examined the transferability of our adversarial mask to a wide range of FR model architectures and datasets.
In addition, we validated our adversarial mask's effectiveness in real-world experiments (CCTV use case) by printing the adversarial pattern on a fabric face mask.
In these experiments, the FR system  was only able to identify 3.34\% of the participants wearing the mask (compared to a minimum of 83.34\% with other evaluated masks).
A demo of our experiments can be found at: \url{https://youtu.be/_TXkDO5z11w}.
\keywords{Adversarial Attack \and Face Recognition \and Face Mask}
\end{abstract}

\section{\label{sec:intro}Introduction}

For the past two years, the coronavirus has impacted every aspect of our lives, and its impact will continue for the foreseeable future.
Since its emergence, various suggestions have been made to reduce its spread.
While the effectiveness of some actions is questionable, there is no doubt that face masks are a key factor in preventing the spread of the virus in crowded and enclosed spaces.
The widespread adoption of face masks and the ever-increasing use of deep learning-based facial recognition (FR) models in everyday systems can be leveraged to perpetrate targeted adversarial attacks that will enable attackers to evade such models and compromise their robustness, without raising an alarm.

Adversarial attacks in the computer vision domain have gained a lot of interest in recent years, and various ways of fooling image classifiers~\cite{szegedy2013intriguing,goodfellow2014explaining} and object detectors~\cite{sitawarin2018darts,thys2019fooling,zolfi2021translucent} have been proposed.
Attacks against FR systems have also been shown to be effective.
For example, research has demonstrated that face synthesis in the digital domain can be used to fool FR models~\cite{yang2020design}.
In the physical domain, some of the proposed methods involved wearing adversarial eyeglasses~\cite{sharif2016accessorize}, projecting lights on human faces~\cite{shen2019vla}, wearing a hat containing an adversarial sticker~\cite{komkov2021advhat}, and using adversarial makeup~\cite{guetta2021dodging}.
However, the proposed attacks are conspicuous and do not allow the attacker to blend in naturally in real-world scenarios, potentially triggering defense systems.

\begin{figure}[t!]
    \centering
    \includegraphics[width=0.7\linewidth]{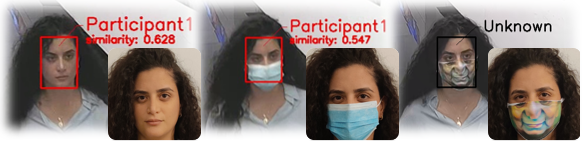}
    \caption{Illustrating the effect of an adversarial pattern printed on a fabric mask (right), which results in the failure of the FR system to detect the person wearing it, compared to the FR system's ability to detect the same individual without a mask, as well as with a standard disposable mask.}
    \label{fig:intro}
\end{figure}

In this work, we propose a \textit{universal} adversarial attack that can be used to physically evade FR systems; in this case, an adversarial pattern is printed on a fabric face mask, as shown in Figure~\ref{fig:intro}.
To create the adversarial pattern, we use a gradient-based optimization process that aims to cause \textit{all} identities wearing the mask to be misclassified by the FR model.
We first demonstrate the attack's ability to fool state-of-the-art models (e.g., ArcFace~\cite{deng2019arcface}) in the digital domain by applying the face mask to every facial image in the dataset (dynamically) using 3D face reconstruction.
Then, we print the adversarial pattern on an actual fabric face mask and test it under real-world conditions.
The results in the digital domain show that our adversarial mask performs better than all evaluated masks and is transferable to other models. 
In the physical domain, we show that 96.66\% of the participants wearing our mask evaded the detection by the FR system.

The contributions of our research can be summarized as follows:
\begin{itemize}
    \item We are the first to present a \textbf{physical universal} adversarial attack that fools FR models, i.e., we craft a single perturbation that causes the FR model to falsely classify all potential attackers as unknown identities, even under diverse conditions (angles, scales, etc.) in a real-world environment (fully-automated CCTV scenario).
    \item In the digital domain, we study the transferability of our attack across different model architectures and datasets.
    \item We present a fully differentiable novel digital masking method that can accurately place any kind of mask on any face, regardless of the position of the head.
    This method can be used for other computer-vision tasks (e.g., training masked-face detection models).
    \item We craft an inconspicuous pattern that ``continues" the contour of the face, allowing a potential attacker to easily blend in with a crowd without raising an alarm, given the variety and widespread use of face masks during the COVID-19 pandemic.
    \item We propose various countermeasures that can be used during the FR model training and inference phases.
\end{itemize}

\section{\label{sec:related}Background \& Related Work}

\subsection{Adversarial Attacks}
% Adversarial attacks against machine learning models have been extensively studied over the past few years, and researchers have proposed various ways of fooling deep neural networks (DNNs) in the computer vision domain. 
% These methods can be roughly categorized as: (a) digital attacks, which operate in the digital space, or (b) physical attacks, which operate in the real world.
% Adversarial attacks in the computer vision domain can be roughly categorized as: digital attacks and physical attacks.

\noindent\textbf{Digital Attacks.}
Initially, attacks in the digital domain aimed at fooling classification models were introduced~\cite{szegedy2013intriguing,goodfellow2014explaining}.
While those earlier attacks are based on methods that generate a perturbation for a single image, Moosavi-Dezfooli \etal~\cite{moosavi2017universal} proposed universal adversarial perturbations (UAPs), which enable any image that is blended with the UAP to fool a DNN.
Digital attacks on models that perform more complex computer vision tasks (e.g., face recognition and object detection) have also emerged.
% Liu~\etal~\cite{liu2018dpatch} presented a patch-based attack on object detectors called \textit{DPatch}. 
% By placing a digital patch in the corner of an image (i.e., not on the targeted object itself), they were able to deceive state-of-the-art object detectors.
Yang~\etal~\cite{yang2020design} designed a digital patch which is placed on a person’s forehead to deceive face detectors.
Recent studies targeting FR models suggested various techniques. 
Deb~\etal~\cite{deb2020advfaces} proposed automated adversarial face synthesis, using a generative adversarial network (GAN) to create minimal perturbations.
Agarwal~\etal~\cite{agarwal2018image} and Amada~\etal~\cite{amada2021universal} proposed UAPs that can deceive FR models for multiple identities simultaneously.
However, these attacks only call attention to the potential threat inherent to such models but cannot be transferred to the physical world.

\noindent\textbf{Physical Attacks.}
% In most of the attacks proposed for the physical domain (i.e., real world), the perturbation is crafted in the digital space, just like attacks in the digital domain.
% However, they differ, because real-world constraints are considered throughout the process of generating the perturbation, and these constraints allow the perturbations to transfer more easily to the physical world.
Physical attacks differ from digital attacks in the way real-world constraints are considered throughout the process of generating the perturbation.
Consequently, these constraints allow the perturbations to transfer more easily to the physical world.
In recent years, physical attacks on object detectors have gained attention.
% Eykholt~\etal~\cite{eykholt2018robust} proposed attaching black and white stickers on a stop sign in a certain pattern to fool image classifiers.
Chen~\etal~\cite{chen2018shapeshifter} printed stop signs containing adversarial patterns that evaded detection by the object detector, and Sitawarin~\etal~\cite{sitawarin2018darts} deceived autonomous car systems by crafting toxic traffic signs that look similar to the original traffic signs.
Methods against person detectors have also been proposed.
Thys~\etal~\cite{thys2019fooling} suggested attaching a small adversarial cardboard plate to a person's body to evade detection.
Continuing this line of research, other studies involved printing adversarial patterns on t-shirts, which resulted in a more realistic article of clothing that blends into the environment more naturally~\cite{wu2020making,xu2020adversarial}.
A slightly different approach, in which the perturbation affects the sensor’s perception of the object by applying a translucent patch on the camera's lens, was also introduced~\cite{zolfi2021translucent}.
% The authors applied a translucent patch on the camera's lens to fool image classifiers.
% Then, Zolfi~\etal~\cite{zolfi2021translucent} improved this technique to fool object detectors for all class instances.

Numerous studies have demonstrated different ways of fooling FR systems.
For example, Shen~\etal~\cite{shen2019vla} introduced the visible light-based attack, where lights are projected on human faces.
Other studies showed that carefully applied makeup patterns can negatively affect the performance of FR systems~\cite{yin2021adv,guetta2021dodging}.
Accessories were also shown to be effective; for example, Sharif~\etal~\cite{sharif2016accessorize} suggested wearing adversarial eyeglass frames that were crafted using gradient-based methods.
Later, GAN methods were used to generate an enhanced version of the adversarial eyeglass frames~\cite{sharif2019general}.
Recently, Komkov~\etal~\cite{komkov2021advhat} printed an adversarial paper sticker and placed it on a hat to fool the state-of-the-art \textit{ArcFace}~\cite{deng2019arcface} FR model.
However, when implemented on a person, these methods may call attention to the person by causing them to stand out in a crowd given their unnatural appearance. 
In contrast, we propose a method in which the perturbation is placed on a face mask, a safety measure widely used in the COVID-19 era; in addition, unlike prior work in which the proposed attacks craft tailor-made perturbations (target a single image or person), our universal attack can be applied more widely without the need for an expert to train a tailor-made one.
Furthermore, we demonstrate the effectiveness of our method in a real-world use case involving a CCTV system, an aspect not addressed by previous studies.

\subsection{Face Recognition}

\textbf{Models.} FR models can be categorized by two main attributes, the model's backbone and the novel loss function, both of which are involved in the training phase.
The main architecture used as the backbone in these models is the ResNet~\cite{he2016deep} architecture, which varies in terms of the number of layers it contains, also referred to as the backbone \textit{depth}.
On top of the backbone, an additional layer (or more) is added, usually containing a novel loss function that is used to train the backbone weights~\cite{deng2019arcface,wang2018cosface,meng2021magface}.
Later, when the FR model is used for inference, only the backbone layers are used to generate the embedding vector.

\noindent \textbf{Systems.} The end-to-end procedure of a fully automated FR system consists of several main steps: (a) Record - a camera records the environment and then produces a series of frames (a video stream); (b) Detect - each frame is analyzed by a face detector to extract cropped faces; (c) Align - the cropped faces are aligned according to the FR model's alignment method; (d) Embed - the aligned facial images serve as input to an FR model $f$ that maps a facial image $I_{face}$ to a vector $f(I_{face})$, also referred to as an \textit{embedding} vector; (e) Verify - the embedding vector is compared to a list of precalculated embedding vectors (also referred to as ground-truth embedding vectors) using a similarity measure (e.g., cosine similarity). The identity with the highest similarity score is marked as a potential candidate and eventually confirmed if its similarity score surpasses a predefined verification threshold (which depends on the system's use case).

\section{\label{sec:method}Method}

The objective of our research is to generate an adversarial pattern that can be printed on a face mask and cause FR systems to classify a registered identity as an unknown identity.
Further, we aim to create an adversarial pattern that is: (a) universal - it must be effective on any identity from multiple views and angles, and at multiple scales, (b) practical - the pattern should remain adversarial when printed on a fabric mask in the real world, and (c) transferable - it must be effective on different models (backbone depths and loss functions). 

\begin{figure}[h]
\centering
    \includegraphics[width=0.7\linewidth]{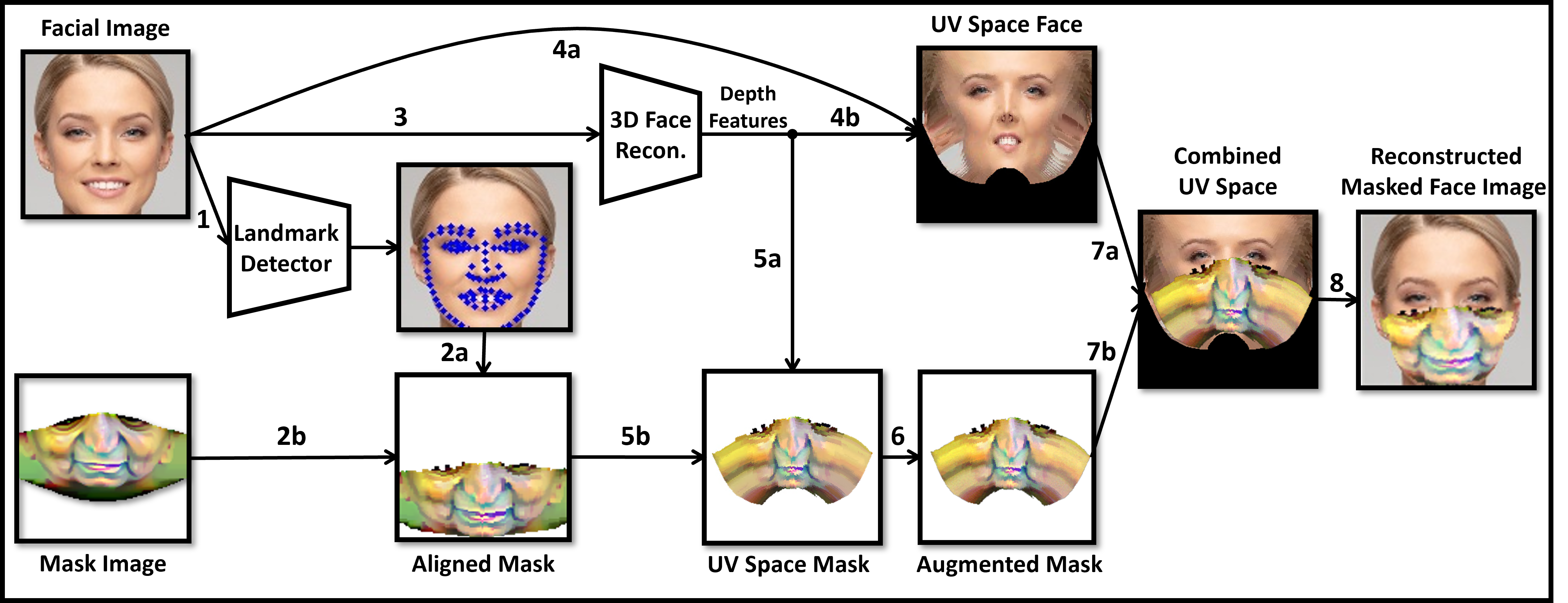}
    \caption{Overview of our mask projection method pipeline.}
    \label{fig:mask_placement}
\end{figure}

\subsection{Mask Projection}

% In this research, we strive to produce an adversarial patch in the form of a face mask.
In order to digitally train our adversarial mask, we first need to simulate the mask overlay on a person's face in the real world.
Therefore, we use 3D face reconstruction to digitally apply a mask on a facial image.
Feng~\etal~\cite{feng2018joint} introduced an end-to-end approach called \textit{UV position map} that records the 3D coordinates of a complete facial point cloud using a 2D image.
This map records the position information of a 3D face and provides dense correspondence to the semantic meaning of each point in the UV space, allowing us to achieve near-real approximation of the mask on the face, which is essential to the creation of a successful adversarial mask in the real world.

More formally, we consider our mask $M_{\text{adv}} \in \mathbb{R}^{w \times h \times 3}$ and a rendering function $\mathcal{R}_\theta$.
The rendering function (partially inspired from~\cite{wang2021facex}) takes a mask $M_{\text{adv}}$ and a facial image $x_{\text{face}}$, and applies the mask on the face, resulting in a masked face image $\mathcal{R}_\theta(M_{\text{adv}}, x_{\text{face}})$.
As shown in Figure~\ref{fig:mask_placement}, the pipeline of the mask's projection on the facial image is as follows:

\begin{enumerate}
    \item Detect the landmark points of the face - given a landmark detector, we extract the landmark points of the face.
    \item Map the mask pixels to the facial image - the landmark points of the face extracted in the previous step of the pipeline are used to map the mask pixels to the corresponding location on the facial images.
    \item Extract depth features of the face - the facial image is passed to the 3D face reconstruction model to obtain depth features.
    \item Transfer 2D facial image to the UV space - the depth features are used to remap the facial image to the UV space.
    \item Transfer 2D mask image to the UV space - the depth features are used to remap the mask image to the UV space.
    \item Augment mask - to improve the robustness of our adversarial mask, random geometric transformations and color-based augmentations (parameterized by $\theta$) are applied: (i) geometric transformations - random translation and rotation are added to simulate possible distortions in the mask's placement on the face in the real world, and (ii) color-based augmentations - random contrast, brightness, and noise are added to simulate changes in the appearance of the mask that might result from various factors (e.g., lighting, noise or blurring caused when the camera captures the image).
    % \noindent These transformations are parameterized by $\theta$.
    \item Combine and reconstruct - the UV representations of the facial image and the mask are combined, and the combined image is reconstructed back to the regular 2D space, resulting in a masked face image.
\end{enumerate}

\noindent Usually, adversarial attacks that employ textile-like objects (e.g., wearable t-shirt~\cite{wu2020making, xu2020adversarial}) use thin plate splines (TPSs)~\cite{bookstein1989principal} to simulate fabric distortions.
In contrast to these studies, although we aim to craft a textile-based mask, in our case, the mask form on the face remains steady and is not subject to significant distortions.
In addition, our 3D approach allows us to simulate smaller distortions (e.g., caused by the nose shape) without actively using TPSs.

Above all, it is important to note that the entire process presented is completely differentiable and allows us to backpropagate and update the mask pixels.

\subsection{\label{subsec:patch_optimization}Patch Optimization}

To optimize our mask's pixels, we propose an iterative optimization process.
In each iteration, we select a random batch of facial images of multiple identities and digitally project the mask on each facial image.
We then feed the masked face images to the FR model and obtain the embedding representations.
Since our goal is to cause an attacker to be unknown to FR models, we aim to create a patch $M_{adv}$ that will decrease the similarity between the output embedding and the ground-truth embedding $e_{gt}$ (precalculated) for each identity.

More formally, an FR model $f:\mathcal{X}^{w\times h\times 3}\to\mathbb{R}^N$ receives a facial image $x\in\mathcal{X}$ (in our case, a masked face image $\mathcal{R}_\theta(M_{adv},x)$) as input and outputs the embedding representation $f(\mathcal{R}_\theta(M_{adv},x))$.
% In our case, the model takes a masked face image $\mathcal{R}_\theta(M_{adv},x)$ as input and outputs an embedding vector.
Therefore, we minimize the cosine similarity between the embedding vectors and use the following loss function:
\begin{equation}
    \ell_{sim}(M_{adv}) = \mathbb{E}_{\theta, x} [\text{cos}(f(\mathcal{R}_\theta(M_{adv},x)), e_{gt})]
    \label{eq:l_sim}
\end{equation}
\noindent Since our method is not system-dependent (i.e., does not use a fixed verification threshold determined by a specific use case), we aim to decrease the similarity to the fullest extent possible, in order to perform the most successful attack.

To improve the mask's transferability to other models, we train our patch using an ensemble of FR models, denoted as $J$.
We replace \ref{eq:l_sim} with the following:
\begin{equation}
    \ell_{sim}(M_{adv}) = \mathbb{E}_{\theta, x} \frac{1}{\vert J\vert}\sum\nolimits_j\text{cos}(f^{(j)}(\mathcal{R}_\theta(M_{adv},x)), e_{gt}^{(j)}),
    \label{eq:l_sim_ens}
\end{equation}
where $f^{(j)}$ denotes the $j^{th}$ model and $e_{gt}^{(j)}$ denotes the embedding representation calculated using the $j^{th}$ model.

We also include the \textit{total variation (TV)}~\cite{sharif2016accessorize} factor to ensure that the optimizer favors smooth color transitions between neighboring pixels and is calculated on the mask pixels as follows:
\begin{equation}
    \ell_{TV} = \sum\nolimits_{i,k}\sqrt{(p_{i,k}-p_{i+1,k})^2 + (p_{i,k}-p_{i,k+1})^2}
\end{equation}
\noindent When neighboring pixels are not similar, the penalty of this component is greater.

To be more precise, since the output of $\ell_{sim}$ is in the range of~$[-1,1]$ and the output of $\ell_{TV}$ is in the range of~$[0, 1]$, we transform $\ell_{sim}$ so it is in the same range ($[0, 1]$); thus, we replace \ref{eq:l_sim_ens} with the following:
\begin{equation}
    \ell_{sim}(M_{adv}) = \mathbb{E}_{\theta, x} \frac{1}{\vert J\vert}\sum\nolimits_j\frac{\text{cos}(f^{(j)}(\mathcal{R}_\theta(M_{adv},x)), e_{gt}^{(j)})+1}{2}
\end{equation}

\noindent Finally, the optimization problem we solve is as follows:
\begin{equation}
    \min\limits_{M_{adv}} [\ell_{sim}(M_{adv}) + \lambda * \ell_{TV}(M_{adv})],
\end{equation}
\noindent where $\lambda$ is set at a low value.

\section{\label{sec:eval}Evaluation}

In our evaluation, we first run experiments in the digital domain by applying the mask to facial images, using the rendering function $R_\theta$ (as explained in Section~\ref{sec:method}).
Then, we evaluate the performance of our adversarial pattern in the physical domain (i.e., real world) by printing it on a fabric mask.

\noindent\textbf{Models.}
We use three different types of loss functions that were originally used to train the models, which are considered state-of-the-art:  ArcFace~\cite{deng2019arcface}, CosFace~\cite{wang2018cosface}, and MagFace~\cite{meng2021magface}.
Specifically, we use pretrained models which were trained using the ArcFace and CosFace loss functions~\cite{an2020partical_fc}, with four different ResNet depths (18, 34, 50, and 100) each, and a pretrained ResNet100 backbone originally trained with the MagFace~\cite{meng2021magface} loss function, for a total of nine different models.
We examine multiple training variations, using one or more (i.e., ensemble) models to train the adversarial mask and then test it in a white-box setting to evaluate the performance. 
We also evaluate the transferability of our mask to other unknown models (i.e., black-box setting).

\noindent\textbf{Datasets.}
Throughout this paper, we use three commonly used datasets in the face recognition domain: CASIA-WebFace~\cite{yi2014learning}, CelebA~\cite{liu2018large}, and MS-Celeb~\cite{guo2016ms}.

For the training phase, we randomly choose 100 different identities (50 men and 50 women) from the CASIA-WebFace dataset.
We extract five random facial images for each identity, for a total of 500 facial images.% for training the adversarial mask.

For the evaluation phase, we use 200 identities from each dataset (an equal number of men and women from each dataset), evaluating both the performance on the same distribution (different identities from the CASIA-WebFace dataset, $\sim$20K images) and the transferability to other datasets (CelebA and MS-Celeb, $\sim$6K and $\sim$24K images, respectively).

\noindent\textbf{Metrics.}
In our experiments, we quantify the performance of our attack as the ability to decrease the similarity score - specifically the cosine similarity (an approach originally presented in ~\cite{komkov2021advhat}).
The cosine similarity calculation is a step required prior to making a binary decision based on a predefined threshold.
This evaluation approach does not require a system-dependent predefined threshold and demonstrates our attack's effectiveness.
In the physical domain, we also quantify the effectiveness of our attack using two additional metrics, each of which relates to a different stage of an end-to-end FR system:
\begin{itemize}
    \item \textit{Recognition rate} (RR) = $|F_{rec}| \mathbin{/} |F_{det}|$,
    where $|F_{rec}|$ denotes the total number of frames in which the identity was correctly recognized (the cosine similarity between the ground-truth embedding and the output embedding surpasses the verification threshold), and $|F_{det}|$ denotes the total number of frames in which a face was detected and analyzed by the FR system.
    \item \textit{Persistence detection} - since the goal of our adversarial mask is to ensure that an attacker is not identified by the system, we propose a metric that indicates whether the goal was met.
    An attacker is considered as identified if, within a window of $N_{\text{sliding window}}$ frames, the attacker was recognized in $N_{\text{recognized}}$ frames (where $N_{\text{recognized}}\leq N_{\text{sliding window}})$.
\end{itemize}

\noindent\textbf{Implementation details.}
The models we work with in this research only take size ${3 \times 112 \times 112}$ facial images as input.
Therefore, We set the size of our patch to be $3 \times 60 \times 112$ to avoid significant downsampling when dynamically rendering the mask to the facial image, and we set the initial color of the mask to white.
The pixels are updated using the Adam optimizer~\cite{kingma2014adam}, where the initial learning rate is set at $10^{-2}$.
The weight factor of the TV component in the loss function $\lambda$ is manually set at $0.1$.
The source code is available online.\footnote{ \url{https://github.com/AlonZolfi/AdversarialMask}}
% Since our implementation is entirely differentiable, we use an automatic differentiation tool kit (PyTorch) to optimize the mask pixels using the backpropagation algorithm.

% To dynamically detect the landmarks of the original faces, we use a lightweight network~\cite{PFL} that is based on the MobileFaceNet~\cite{chen2018mobilefacenets} architecture.
% We chose a small network to keep our end-to-end masking process fast yet dynamic.

% \blfootnote{The source code can be found at: github link will be added in the final version.}

\begin{figure}[t!]
    \centering
    \begin{subfigure}{0.15\linewidth}
        \centering
        \includegraphics[width=0.97\linewidth]{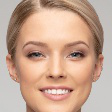}
        \caption{Clean}
    \end{subfigure}
    \begin{subfigure}{0.15\linewidth}
        \centering
        \includegraphics[width=0.97\linewidth]{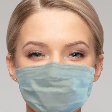}
        \caption{Blue}
        \label{subfig:digital_blue}
    \end{subfigure}
    \begin{subfigure}{0.15\linewidth}
        \centering
        \includegraphics[width=0.97\linewidth]{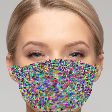}
        \caption{Random}
    \end{subfigure}
    \begin{subfigure}{0.15\linewidth}
        \centering
        \includegraphics[width=0.97\linewidth]{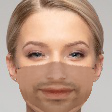}
        \caption{Male}
    \end{subfigure}
    \begin{subfigure}{0.15\linewidth}
        \centering
        \includegraphics[width=0.97\linewidth]{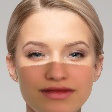}
        \caption{Female}
    \end{subfigure}
    \begin{subfigure}{0.15\linewidth}
        \centering
        \includegraphics[width=0.97\linewidth]{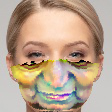}
        \caption{Adv}
    \end{subfigure}
    \caption{Examples of facial images w/o mask (a), and when various masks are digitally applied to them (b)-(f).}
    \label{fig:masks_examples}
\end{figure}

\noindent\textbf{Types of face masks evaluated.}
Since we are the first to present a physical universal perturbation, we compare the effectiveness of our mask with several control masks: (a) Clean - the original facial image without a mask, (b) Adv - our optimized adversarial mask, (c) Random - a mask with randomly colored pixels, and (d) Blue - a standard disposable blue mask (simple black and white masks were also tested and yielded the same results).
In addition, due to our trained mask's resemblance to a human face, the lower face area of a female and male are used as control masks and will be referred to as \textit{Female Face} and \textit{Male Face}, respectively.
The masks compared in our evaluation are shown in Figure~\ref{fig:masks_examples}.

\noindent\textbf{Evaluation setup.}
Since the state-of-the-art models discussed above were not specifically designed to address the issue of masked faces, we first examine the model's (ResNet100@ArcFace) performance on a number of simple face masks.
For this evaluation, we use 100 identities from the CASIA-WebFace dataset, where five images of each identity are used to calculate the ground-truth embedding, and the remaining images are applied with different types of masks.

To the best of our knowledge, the scientific community has not reached a consensus on the way in which masked face images should be dealt with by FR models.
Therefore, we use two approaches for generating the ground-truth embedding: (a) the current approach for unmasked face models - averaging the embedding vectors of the original images only, and (b) an extension of the first approach - in addition to the original images, we create a masked face version for each image (the specific mask is randomly chosen from blue, black, and white masks) and average the embedding vectors of the two versions of the images.
We then calculate the cosine similarity between the masked face images' embedding vectors and the two versions of ground-truth embedding vectors generation.

In Table~\ref{tab:mask_performance} we can see that although the first approach (w/o Mask) performs better on unmasked images, its performance on masked images is unsatisfactory.
On the other hand, the cosine similarity for the second approach (w/Mask) only slightly decreases the cosine similarity on unmasked images ($\sim$0.05 decrease) and performs significantly better on masked images ($\sim$0.1-0.15 increase).
Thus, throughout this section the results we present are obtained using the second approach (the ground-truth embedding vectors used for the training procedure are generated using first approach).
It is important to note that by choosing the second approach, we increase the difficulty of deceiving these models, since the ground-truth embedding vectors encapsulate the use of a face mask.

\begin{table}[t!]
\centering
\captionof{table}{Cosine similarity comparison between two ground-truth embedding generation methods on the Resnet100@ArcFace. Bold indicates better performance.}

\begin{tabular}{c|c|cccc}
\hline
                                                                             & \textbf{Mask Type} & No Mask       & Blue          & Black         & White         \\ \hline \hline
\multirow{2}{*}{\textbf{\begin{tabular}[c]{@{}c@{}}Cosine\\ Similarity\end{tabular}}} & w/o Mask*           & \textbf{.732} & .399          & .407          & .428          \\ \cline{2-2}
                                                                             & w/Mask**           & .682          & \textbf{.547} & \textbf{.549} & \textbf{.561} \\ \hline
\end{tabular}
\caption*{
\scriptsize
*Embedding vectors created using original facial images.\\
**A masked version of the original images is added to the embedding calculation.}
\label{tab:mask_performance}
\end{table}

\subsection{Digital Attacks}

We conduct digital experiments to quantify our adversarial mask's effectiveness using the rendering function $R_\theta$ (see Section~\ref{sec:method}), which allows us to dynamically apply masks to the facial images in the test set.

\begin{figure}[t!]
\centering
    \includegraphics[width=0.6\linewidth]{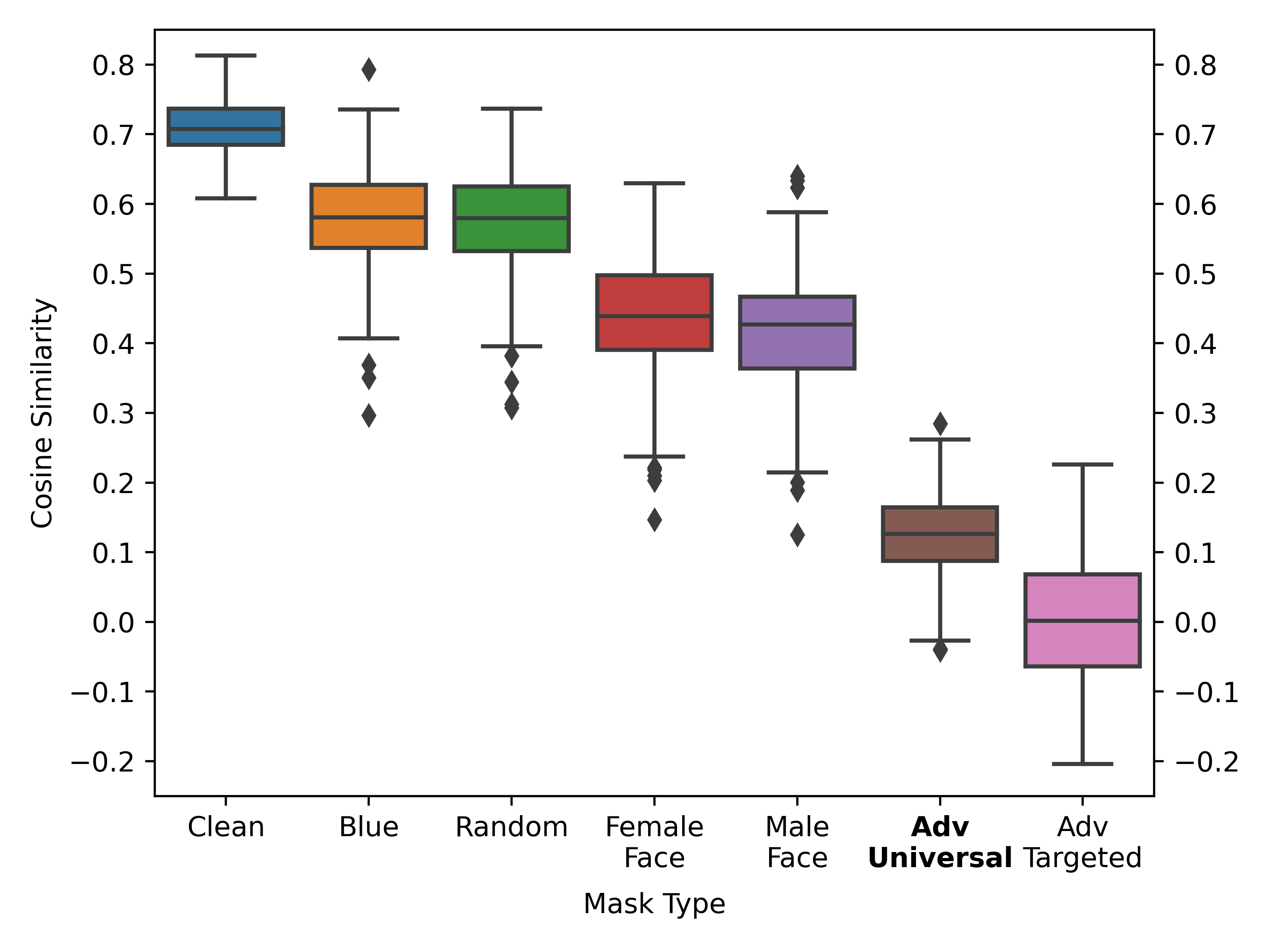}
    \caption{Distribution of the cosine similarity score across different masks. 'Adv Universal' represents our optimized universal mask, and 'Adv Targeted' represents a tailor-made mask for each identity.}
    \label{fig:simbox}
\end{figure}

\noindent\textbf{Effectiveness of the adversarial mask in a white-box setting.}
We examine the effectiveness of our attack in a \textit{white-box} setting in which our mask is optimized and tested on the ResNet100@ArcFace. 
As shown in Figure~\ref{fig:simbox}, our adversarial mask has a significant impact compared to the no mask case, in which the average cosine similarity decreased from $\sim0.7$ to $\sim0.1$.
As the case of no mask images represents the upper bound of the cosine similarity, we also perform a targeted attack in which a mask is tailored to each person, to determine the lower bound.
The targeted mask results are averaged across all identities in the test set.
We can see that the universal mask performs almost as good as a tailor-made mask ($\sim0.1$ difference).
The tailor-made masks represent an attack that is more difficult to detect, since the adversarial pattern varies among different identities.
In addition, while the female and male face control masks are also able to decrease the cosine similarity to a lower level ($\sim0.45$), our mask outperforms them both for almost all tested identities.

\begin{figure}[t!]
    \centering
   \label{fig:transferability}
    \begin{subfigure}{0.48\linewidth}
        \centering
        \includegraphics[width=0.85\linewidth]{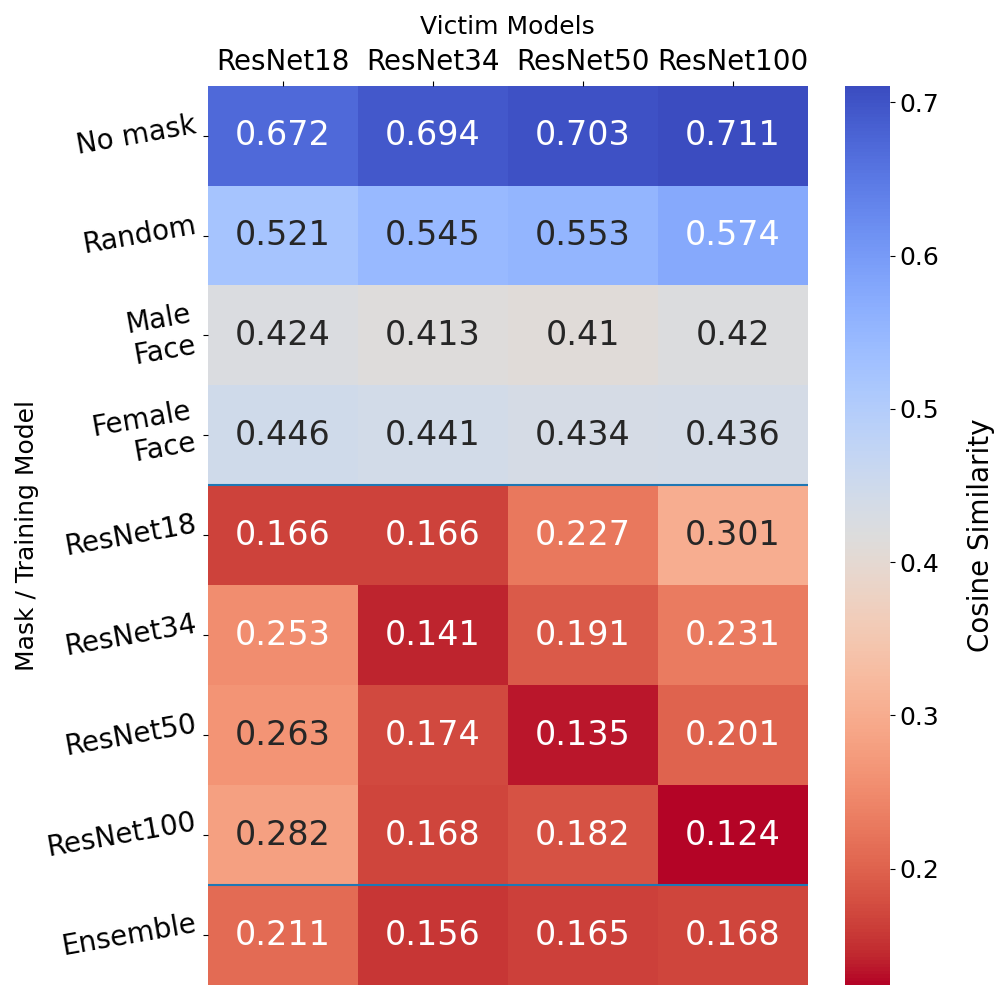}
        \caption{\textbf{Transferability} across various ResNet backbone \textbf{depths} originally trained using the ArcFace loss function.}
        \label{subfig:depth_heatmap}
    \end{subfigure}
    \hspace{8pt}
    \begin{subfigure}{0.48\linewidth}
        \centering
        \includegraphics[width=0.85\linewidth]{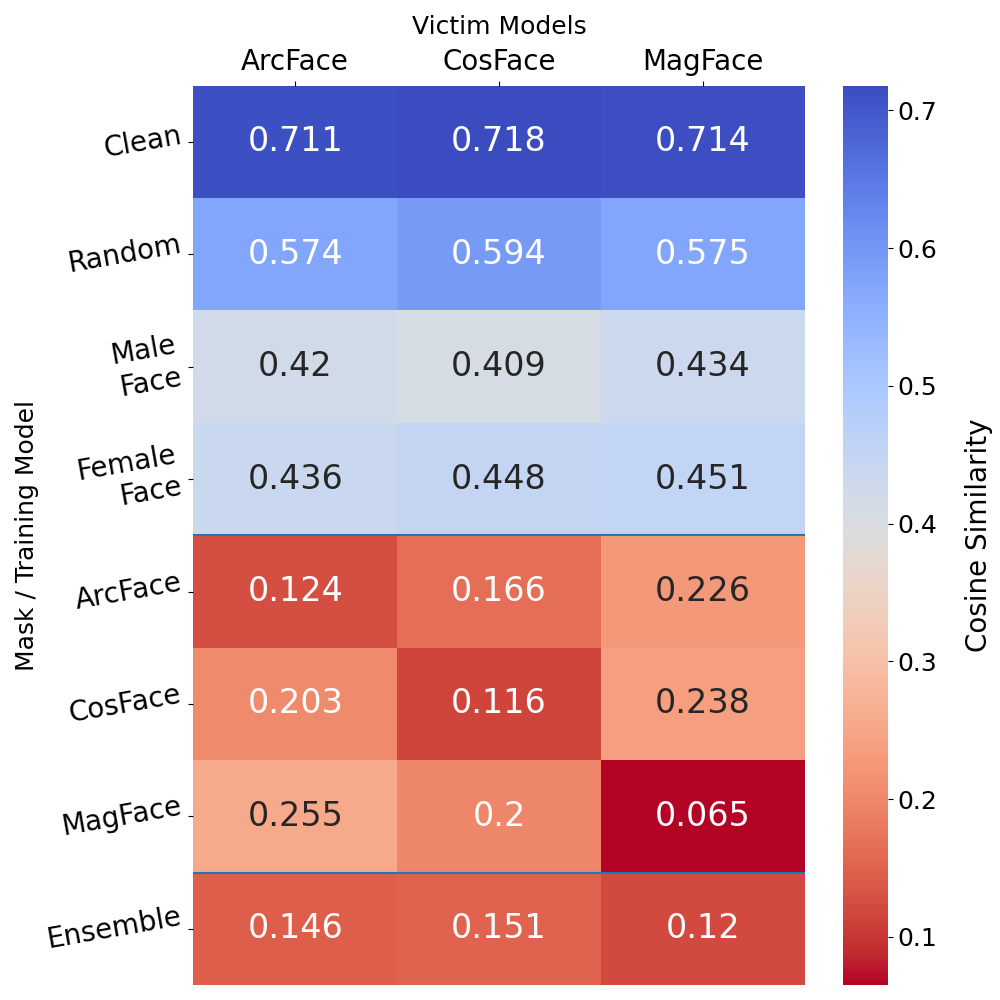}        
        \caption{\textbf{Transferability} across various ResNet100 backbones originally trained with different \textbf{loss functions}.}
        \label{subfig:head_heatmap}
    \end{subfigure}
    \caption{Transferability experiments measured in terms of cosine similarity.
    Rows are divided into three groups: control masks, masks trained using a single model, mask trained using all of the models.}
\end{figure}

\noindent\textbf{Transferability across backbone depth.}
We also examine whether our mask can deceive FR models it was not trained on.
Since the majority of the models use the ResNet architecture, we evaluate the performance across different depths of the ResNet@ArcFace.
The results are presented in Figure~\ref{subfig:depth_heatmap}. 
In the figure, we can see that the use of our adversarial mask can cause the cosine similarity to decrease regardless of the model used for training.
It can also be seen that our attack generalizes better to unknown models whose architecture depth is closer to that of the trained model.
For example, an adversarial mask trained on a model with 100 layers performs better on the models with 34 and 50 layers (decreasing the cosine similarity to 0.182 and 0.168, respectively) than on the 18-layer model (0.282).
In addition, we see that the mask trained on an ensemble of all models does not outperform a mask trained on a single model in a white-box setting, however the ensemble's effectiveness is seen over all models combined.

\begin{figure}[t!]
\centering
    \begin{subfigure}{0.18\linewidth}
        \centering
        \fbox{\includegraphics[width=0.94\linewidth,trim={0 0.8cm 0 0.8cm},clip]{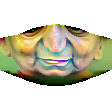}}
        \caption{ArcFace}
        \label{subfig:arcface100_mask}
    \end{subfigure}
    \hspace{0.3cm}
    \begin{subfigure}{0.18\linewidth}
        \centering
        \fbox{\includegraphics[width=0.94\linewidth,trim={0 0.8cm 0 0.8cm},clip]{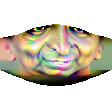}}
        \caption{CosFace}
        \label{subfig:cosface_head}
    \end{subfigure}
    \hspace{0.3cm}
    \begin{subfigure}{0.18\linewidth}
        \centering
        \fbox{\includegraphics[width=0.94\linewidth,trim={0 0.8cm 0 0.8cm},clip]{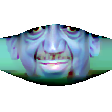}}   
        \caption{MagFace}
        \label{subfig:magface_head}
    \end{subfigure}
    \caption{Illustrations of our adversarial masks trained on different ResNet100 backbones, which vary in terms of the original loss function they were trained on.}
    \label{fig:head_masks}
\end{figure}

\noindent\textbf{Transferability across different loss functions.}
We further demonstrate the adversarial mask's transferability across different model loss functions.
We use the ResNet100 backbone in which the weights were trained using one of the following loss functions: ArcFace, CosFace, and MagFace.
In Figure~\ref{subfig:head_heatmap}, we observe that our method is loss-agnostic, as the decrease in the cosine similarity is seen on for all tested models.
However, a mask that was trained using the MagFace model does not generalize as well as the masks trained with other models, where the cosine similarity decreased to 0.065 in the white-box setting but only decreased to 0.255 and 0.2 on the ArcFace and CosFace models, respectively.
It is interesting to examine the mask trained by each model (presented in Figure~\ref{fig:head_masks}).
Whereas there is a resemblance in the contour of the optimized masks, the mask trained using the ResNet100@MagFace backbone (Figure~\ref{subfig:magface_head}) learns completely different colors than the other two, in some way providing a possible explanation for its decreased ability to generalize to the ArcFace and CosFace models.

\noindent\textbf{Transferability across datasets.}
We also find our mask to be effective across different datasets.
In another experiment, we train our mask using images from one of the examined datasets (presented earlier in this section) and study its effectiveness on the other datasets (i.e., the ground-truth embedding vectors are generated using another dataset's images).
We train all of the masks using the ResNet100@ArcFace.
The results show that the impact of using a specific dataset is insignificant, since our mask generalizes over all datasets.
For example, when training the mask on the CASIA-WebFace dataset and testing it on the CelebA and MS-Celeb datasets, we respectively obtained an average cosine similarity of 0.128 and 0.114, similar to the white-box setting results (mask trained and tested on images from the CASIA-WebFace dataset, Figure~\ref{fig:simbox}).
% As shown in Table~\ref{tab:dataset_trans}, the use of a specific dataset is insignificant, since our mask generalizes over all datasets.

% \begin{table}[t!]
% \centering
% \begin{tabular}{lccc}
% \hline
%          & CASIA & CelebA & MS-Celeb \\ \hline \hline
% CASIA    & 0.106  & 0.128  & 0.114    \\
% CelebA   & 0.107 & 0.095   & 0.103    \\
% MS-Celeb & 0.118 & 0.116  & 0.106   \\ \hline
% \end{tabular}
% \caption{Cosine similarity score when training the mask using images from one dataset and testing on embedding vectors generated using images from other datasets.}
% \label{tab:dataset_trans}
% \end{table}

\begin{figure}[t!]
    \centering
    \begin{subfigure}{0.18\linewidth}
        \fbox{\includegraphics[width=0.95\linewidth,trim={0.1cm 0.9cm 0.1cm 0.9cm},clip]{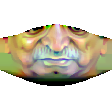}}
        \caption{Female IDs}
        \label{subfig:female}
    \end{subfigure}
    \hspace{0.3cm}
    \begin{subfigure}{0.18\linewidth}
        \fbox{\includegraphics[width=0.95\linewidth,trim={0.1cm 0.9cm 0.1cm 0.9cm},clip]{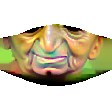}}
        \caption{Male IDs}
        \label{subfig:male}
    \end{subfigure}
    \caption{The adversarial masks trained on the ResNet100@ArcFace using single gender identities.}
    \label{fig:gender}
\end{figure}

\noindent\textbf{Effect of gender.}
Another aspect we studied is the effect of a specific gender on the trained mask.
The experiments include optimization of the adversarial mask using only female or male identities, and the final masks are presented in Figure~\ref{subfig:female} and Figure~\ref{subfig:male}, respectively.
The results show that even when training the mask on facial images of a single gender, the cosine similarity decreases to the same level as the mask trained on both genders ($\sim0.1$).
In addition, masks trained by a single gender were able to transfer very well to the other gender ($\text{male}\to\text{female}=0.097$, and $\text{female}\to\text{male}=0.145$).

Generally, the contour of the trained masks (including the mask trained on both genders, Figure~\ref{subfig:arcface100_mask}) is quite interesting.
Despite the fact that only facial images of female identities were used to train the mask (Figure~\ref{subfig:female}), the optimized mask has an high resemblance to a male face.
More generally, the resemblance of all the trained masks to a male face might indicate there is an underlying bias hidden in these models.

\subsection{Physical Attacks}

\begin{figure}[t!]
    \centering
    \begin{subfigure}{0.18\linewidth}
        \fbox{\includegraphics[width=0.88\linewidth,trim={0.1cm 0.9cm 0.1cm 0.9cm},clip]{figures/arcface100.png}}
        \caption{}
        \label{subfig:surface_digital}
    \end{subfigure}
    \hspace{0.1cm}
    \begin{subfigure}{0.18\linewidth}
        \includegraphics[width=0.98\linewidth]{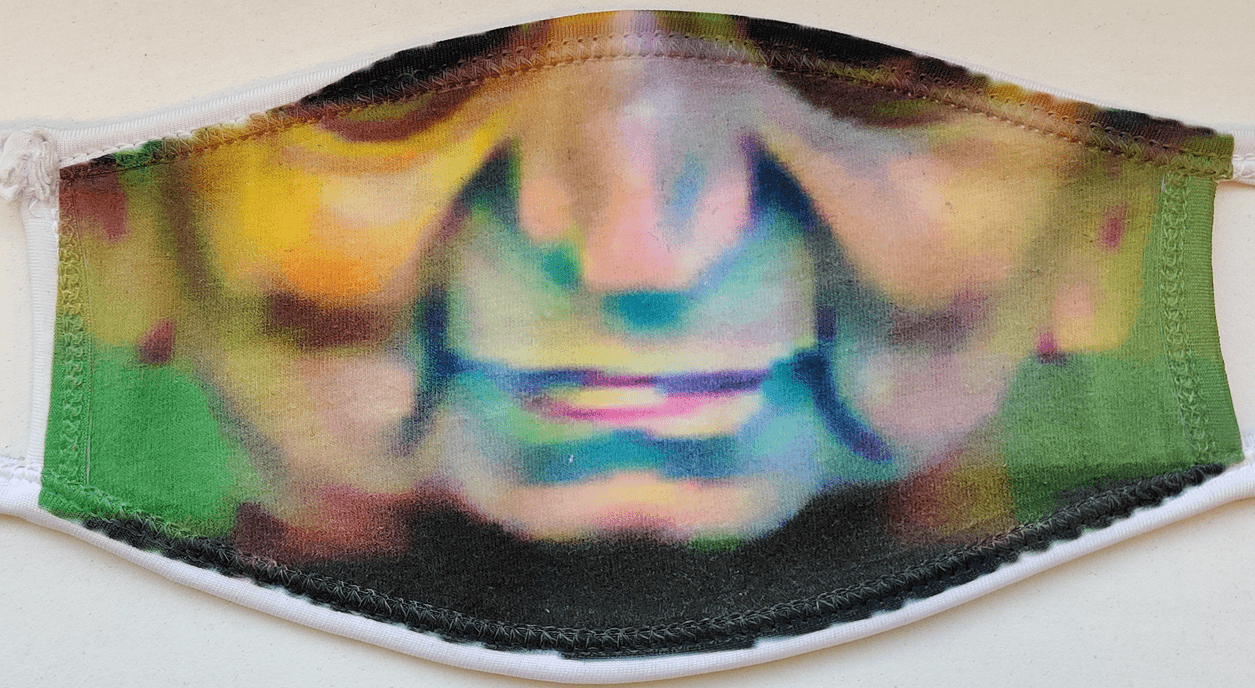}
        \caption{}
        \label{subfig:surface_fabric}
    \end{subfigure}
    \hspace{0.1cm}
    \begin{subfigure}{0.18\linewidth}
        \includegraphics[width=0.98\linewidth]{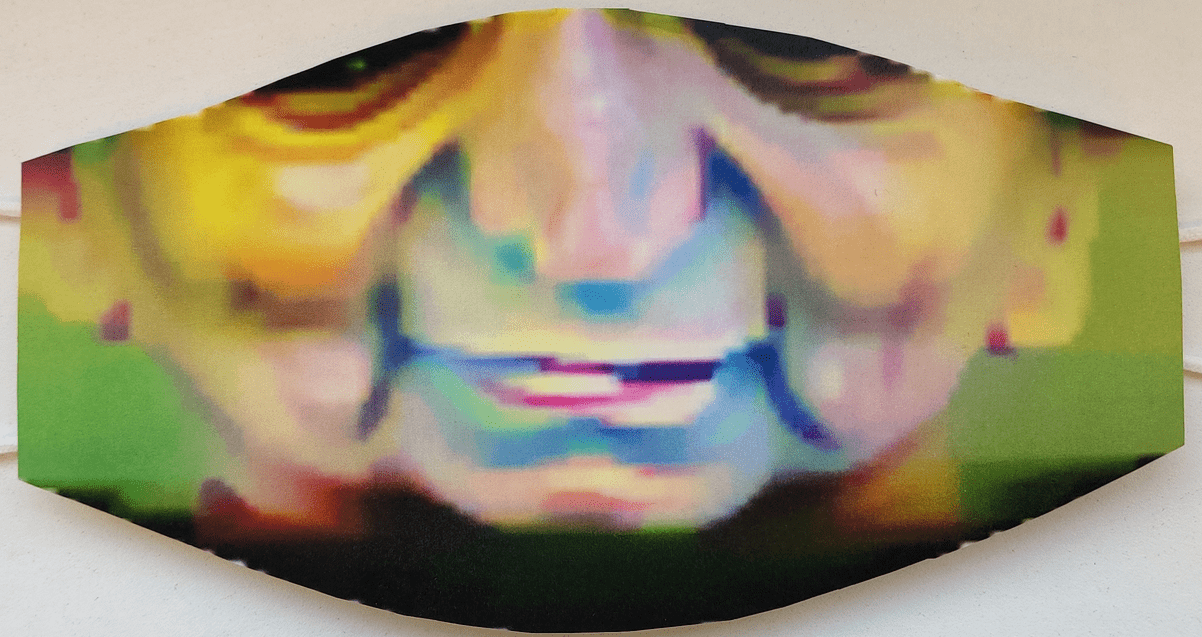}
        \caption{}
        \label{subfig:surface_paper}
    \end{subfigure}
    \caption{An illustration of: (a) the digital adversarial mask trained on the ResNet100@ArcFace; (b) the digital pattern printed on fabric mask; and (c) the digital pattern printed on paper.}
    \label{fig:physical_masks}
\end{figure}

Finally, to evaluate the effectiveness of our attack in the real world, we print our digital pattern on two surfaces: on regular paper cut in the shape of a face mask and on a white fabric mask, as shown in Figure~\ref{fig:physical_masks}.
In addition, we create a testbed that operates an end-to-end fully automated FR system (explained in Section~\ref{sec:related}), simulating a CCTV use case.

\noindent \textbf{Setup.} The system contains: (a) a \textit{Dahua IPC-HDBW1431E} network camera which records a long corridor, (b) an MTCNN~\cite{zhang2016joint} detection model for face detection, preprocessing, and alignment, and (c) an attacked model - we perform a white-box attack in which the model used for training the adversarial mask is also the model under attack, a ResNet100@ArcFace.
In addition, we perform an ``offline" analysis in a black-box setting, in which the facial images are cropped from the original frames and compared to ground-truth embedding vectors generated using other models.
% We use cosine similarity as the similarity metric.

To calculate the specific verification threshold (set at 0.38), we use a subset of 1,000 identities from the CASIA-WebFace dataset and perform the following procedure.
Various face masks are applied (digitally) to each identity's original facial images.
Then, we calculate the cosine similarity between the identity's embedding vector and each masked face image.
Since we employ a semi-critical security use case (CCTV), we chose the threshold that led to a false acceptance rate (FAR) of 1\%.
Furthermore, to minimize false positive alarms, we used a persistence threshold of $N_{\text{recognized}}=7$ frames and a sliding window of $N_{\text{sliding window}}=10$ frames to designate a candidate identity as a valid one.

We recruited a group of 15 male and 15 female participants (after approval was granted by the university's ethics committee).
Each participant was asked to walk along the corridor seven times, once with each mask evaluated (clean, blue, random, male face, and female face), similar to the digital experiments, and two more times with our adversarial masks printed on paper and fabric.
The ground-truth embedding of each participant was calculated using two facial images, where a standard face mask was applied (digitally) to each image, for a total of four facial images.

\begin{figure}[t!]
\centering
    \includegraphics[width=0.7\linewidth]{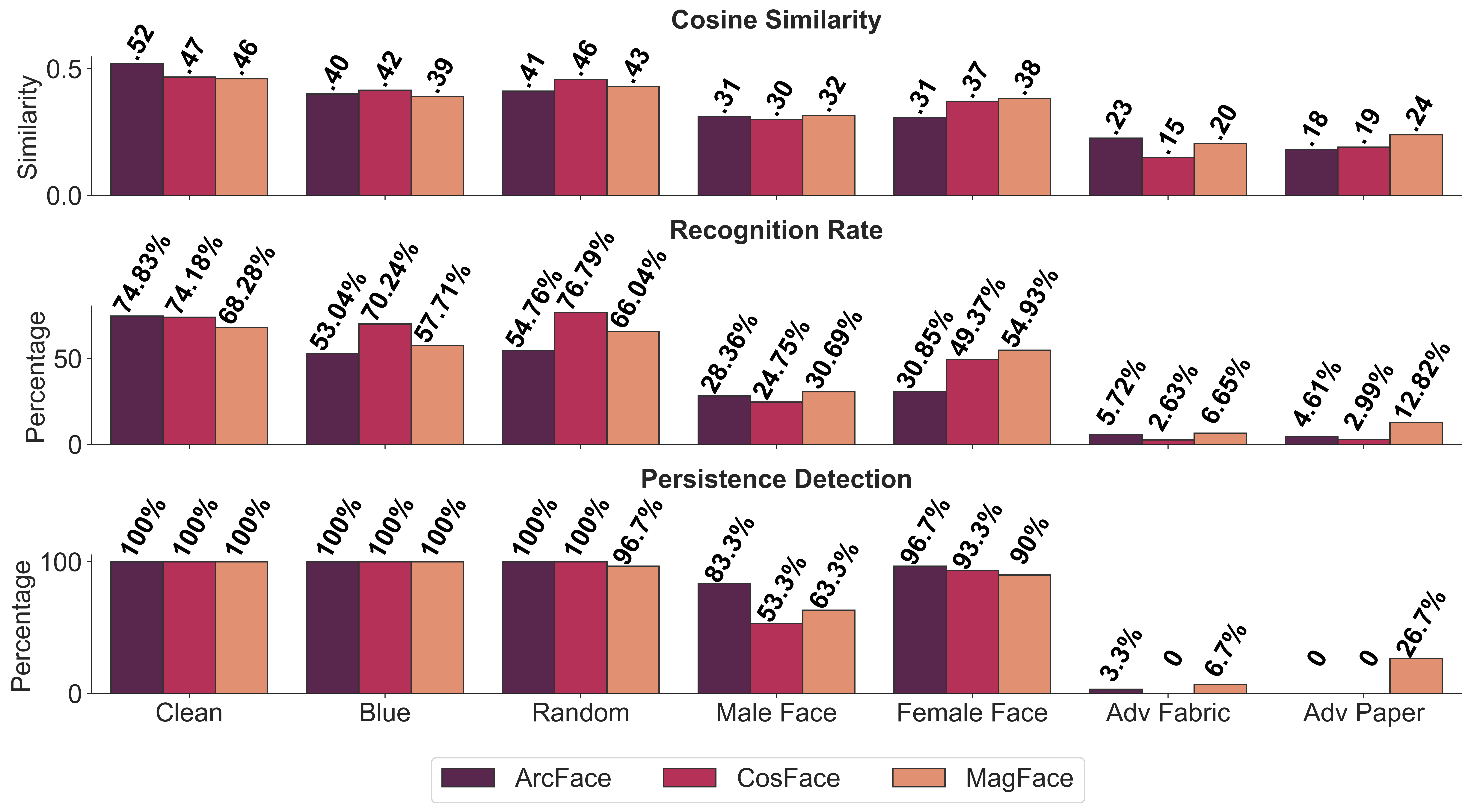}
    \caption{Physical experiments' averaged results on all participants across different evaluated masks and different victim models.}
    \label{fig:physical_results}
\end{figure}

\noindent \textbf{Results.} The results of our experiments are shown in Figure~\ref{fig:physical_results} where we can see that our adversarial masks (paper and fabric) performed significantly better than the other masks evaluated on every metric, with a high correlation to the cosine similarity results obtained in the digital domain.

In terms of the RR, the performance of the FR model for the different masks can be divided into four groups (listed in decreasing order): (a) the unmasked version (74.83\%), (b) blue and random masks (53.04\% and 54.76\%, respectively), (c) male and female masks (30.85\% and 28.36\%, respectively), and (d) our fabric and paper adversarial masks (5.72\% and 4.61\%, respectively).

In a realistic case of CCTV use in which an attacker tries to evade the detection of the system, our adversarial fabric mask was able to conceal the identity of 29 out of 30 participants (which represents a persistence detection value of 3.34\%), as opposed to the control masks which were able to conceal 5 out of 30 participants at most (persistence detection value of 83.34\%).

We also examine the effectiveness of our masks on models they were not trained on.
The results presented in Figure~\ref{fig:physical_results} show that our masks have similar adversarial effect on FR models in a black-box setting as in a white-box setting.
% As explained earlier, the decreased performance on the ResNet100@MagFace could be explained 
% We can also observe that there is a slight difference in the performance on the ResNet100@MagFace.
% Figure~\ref{fig:head_masks} shows that there is a significant visual difference between the ArcFace and MagFace learned masks, which might explain the difference in the performance.

Another aspect we examined in our physical evaluation is the ability to print the adversarial pattern on a real surface.
Figures~\ref{subfig:surface_fabric} and \ref{subfig:surface_paper} present the digital adversarial pattern (\ref{subfig:surface_digital}) printed on the different surfaces.
Due to the limited ability of a printer to accurately output the original colors onto the fabric, we can see that there is a slight difference in the performance of the masks.
Nonetheless, both of our adversarial masks outperformed the other masks evaluated.

\section{Countermeasures}
We propose two ways in which our digital masking method can be used to defend against adversarial masks: (a) adversarial training -- adversarial (universal and tailor-made) masked face images could be provided to the model during training to improve its robustness; and (b) mask substitution -- during the inference phase, every masked face image could be preprocessed so that the worn mask is replaced digitally with a standard one (e.g., blue mask~\ref{subfig:digital_blue}), where the models had satisfactory performance, as shown in Section~\ref{sec:eval}, eliminating the potential threat of an adversarial face mask.
An implementation of the mask substitution method on facial images of 100 identities ($\sim10K$ images) from the CASIA-WebFace dataset increased the RR from 0.4\% (the adversarial mask is applied to the facial images) to 65.5\% (the blue mask is applied to the adversarial images).
In a physical experiment, in which the blue mask was digitally placed on facial images extracted from the videos frames (videos of participants wearing the adversarial mask), the RR increased from 5.72\% to 57.3\%.

\section{\label{sec:conclusion}Conclusion}
In this paper, we presented a physical universal attack in the form of a face mask against FR systems.
Whereas other attack methods used different accessories that are more conspicuous and do not blend naturally in the environment, our mask will not raise any suspicion due to the widespread use of face masks during the COVID-19 pandemic.
We demonstrated the effectiveness of our mask in the digital domain, both under white-box and black-box settings.
In the physical domain, we showed how our mask is able to prevent the detection of multiple participants in a CCTV use case system.
Moreover, we proposed possible countermeasures to deal with such attacks. 
To sum up, in this research, we highlight the potential risk FR models face from an adversary simply wearing a carefully crafted adversarial face mask in the COVID-19 era.

\bibliographystyle{splncs04}
\bibliography{egbib}

\end{document}